\documentclass[a4paper]{article}

\usepackage{INTERSPEECH2018}
\usepackage{times}
\usepackage{latexsym}

\usepackage{array, caption, tabularx, makecell, booktabs}%
\usepackage{multirow}
\usepackage{color}
\usepackage{colortbl}
\usepackage[colorlinks]{}
\usepackage{subfig}
\usepackage{wrapfig}
\usepackage{graphicx}
\usepackage{url}
\usepackage [autostyle, english = american]{csquotes}
\MakeOuterQuote{"}
\usepackage{fixltx2e}

\newcolumntype{P}[1]{>{\centering\arraybackslash}p{#1}}

\title{Neural MultiVoice Models for Expressing Novel Personalities in Dialog}
\name{Shereen Oraby, Lena Reed, Sharath TS, Shubhangi Tandon, Marilyn Walker}
\address{
  Natural Language and Dialog Systems Lab, University of California, Santa Cruz
  }
\email{\{soraby,shtandon,struvek,lireed,mawalker\}@ucsc.edu}

\begin{document}

\maketitle
\begin{abstract}
Natural language generators for task-oriented dialog should be able
to vary the style of the output utterance while still effectively
realizing the system dialog actions and their associated semantics.
While the use of neural generation for training the response
generation component of conversational agents promises to simplify the
process of producing high quality responses in new domains, to our
knowledge, there has been very little investigation of neural
generators for task-oriented dialog that can vary their response
style, and we know of no experiments on models that can generate
responses that are different in style from those seen during training,
while still maintaining semantic fidelity to the input meaning
representation.  Here, we show that a model that is trained to achieve
a single stylistic personality target can produce outputs that combine
stylistic targets. We carefully evaluate the multivoice outputs for
both semantic fidelity and for similarities to and differences from
the linguistic features that characterize the original training
style. We show that contrary to our predictions, the learned models
do not always simply interpolate model parameters, but rather produce
styles that are distinct, and novel from the personalities
they were trained on.
\end{abstract}

\section{Introduction}

Natural language generators for task-oriented dialog should be able
to vary the style of the output while still effectively
realizing the system dialog actions and their associated semantics.
The use of neural natural language generation ({\sc nnlg}) for
training the response generation component of conversational agents
promises to simplify the process of producing high quality responses
in new domains by relying on the neural architecture to automatically
learn how to map an input meaning representation to an output
utterance.
However, there has been little investigation of
{\sc nnlg}s for dialog that can vary
their response style, and we know of no experiments on models that can
generate responses that are different in style from those seen during
training, while still maintaining semantic fidelity to the input
meaning representation. Instead, work on stylistic transfer has
focused on tasks where only coarse-grained semantic fidelity is
needed, such as controlling the sentiment of the utterance (positive
or negative), or the topic or entity under discussion 
\cite{sennrich2016controlling,shen2017style,fancontrollable}. 

Consider for example a training instance for the restaurant domain
consisting of a meaning representation (MR) from the End-to-End (E2E) Generation
Challenge\footnote{\url{http://www.macs.hw.ac.uk/InteractionLab/E2E/}} and a sample output from one of our neural generation models in
Figure~\ref{table:MR} \cite{novikova2016crowd,oraby2018style}.  Systems
using the training set of 50K crowdsourced utterances from the E2E
task achieved high semantic correctness, e.g. the BLEU score for our
best system  on the dev set was 0.72 \cite{novikova2017e2e}.
However in the best case these models can only reproduce the style of
the training data, and in actuality the outputs have reduced stylistic
variation, because when particular stylistic variations are less frequent, they
are treated similarly to noise. 

\begin{table}[t!bh]
\begin{footnotesize}
\begin{tabular}
{@{} p{2.95in} @{}} \toprule
\sc 
inform(name[Browns Cambridge], eatType[pub],  priceRange[average],   food[Italian],
near[Adriatic])  familyFriendly[yes], area[city centre] \\ \hline
{\it Browns Cambridge is a pub, also it is a moderately priced italian place near Adriatic, also it is family friendly, you know and it's in the city centre.}   \\ \hline
\end{tabular}
 \caption{Meaning representation and output training pair \label{table:MR}}
\end{footnotesize}
\end{table}

In subsequent work, we showed that we could augment the E2E training
data with synthetically generated stylistic variants and train a
neural generator to reproduce these variants, however the models can
still only generate what they have seen in training
\cite{oraby2018style}.  Here, instead, we explore whether a model that is
trained to achieve a single stylistic personality target can produce
outputs that combine stylistic targets, to yield a novel style that is
significantly different than what was seen in training,
while still maintaining high semantic correctness. We
first train each stylistic model with a single latent variable for
supervision, for five different personality models, or voices, based on the Big
Five theory of personality, namely the personality trait styles of
{\sc extravert}, {\sc agreeable}, {\sc disagreeable}, {\sc
  conscientious}, and {\sc unconscientious}. Then, at generation time,
we provide the model with combinations of the stylistic variables,
i.e.  we instruct the {\sc nnlg} to generate multivoice outputs that
combine {\sc extravert} with {\sc disagreeable}, where such combined
outputs never occurred in the training data.

We first describe how we set up our dataset and neural models in
Section~\ref{data-model-sec}, and then present our results in
Section~\ref{results-sec}.  We evaluate the multivoice outputs for
both semantic fidelity and for similarities to and differences from
the linguistic features that characterize the original training
style. We hypothesize that controlling multiple stylistic parameters
is more difficult and will lead to more semantic errors, so we examine
in detail the interaction of stylistic variation and semantic fidelity,
as well as quantifying stylistic fidelity. We leave a discussion of
related work until Section~\ref{rel-conc-sec} where we also conclude.

\begin{table*}[t!]
\centering \caption{\label{table:multi-voice-examples} {MultiVoice generation output 
and comparable singlevoice outputs for  {\sc Disagreeable}, {\sc Extravert} and {\sc Conscientious}
for the meaning representation in Figure~\ref{table:MR}. We count the frequency of periods (Period Agg.) and expletives (Explet. Prag) for multivoice models that utilize {\sc disagreeable})}.}
\begin{scriptsize}
\begin{tabular}
{@{} p{0.1in}|p{0.4in}|p{0.35in}|p{0.35in}|p{4.5in}@{}}
\toprule
 & {\bf \scriptsize  Persona} & {\bf \scriptsize Period Agg.} & {\bf \scriptsize Explet. Prag.} & {\bf \scriptsize  Example} \\ \hline          
\rowcolor [gray]{0.9}1 & DISAG & 5.71 & 2.26 &  Browns Cambridge is damn moderately priced, also it's in city centre. It is a pub. It is an italian place. It is near Adriatic. It is damn family friendly. \\ \hline\hline              
2 & CONSC            & 0.60  & 0.02 & Let's see what we can find on Browns Cambridge. I see, well it is a pub, also it is moderately priced, an italian restaurant near Adriatic and family friendly in city centre. \\\hline
\rowcolor [gray]{0.9}3 & DISAG+\newline CONSC   & 3.81 & 0.84 & Browns Cambridge is an italian place and moderately priced. It is near Adriatic. It is kid friendly. It is a pub. It is in city centre. \\\hline\hline
4 & EXTRA            & 0.18 & 0.01 & Browns Cambridge is a pub, also it is moderately priced, an italian place near Adriatic, also it is family friendly, you know and it's in city centre, you know! \\\hline
\rowcolor [gray]{0.9}5 & DISAG+\newline EXTRA   & 8.07 & 1.92 & Damn moderately priced Browns Cambridge is it 's a pub, also it is near Adriatic. It is an italian place and moderately priced. It is in city centre.  \\\hline\hline
6 & UNCON          & 0.40  & 2.88 & Oh god yeah, i don't know. Browns Cambridge is a pub, also it is damn family friendly, also it's an italian place near Adriatic, also it is darn moderately priced in city centre. \\\hline
\rowcolor [gray]{0.9}7 & DISAG+\newline UNCON & 2.88 & 3.16 & Oh god i mean, i thought everybody knew that Browns Cambridge is a pub, also it is near Adriatic. It is an italian place and moderately priced. It is in city centre.  \\
\bottomrule 
\end{tabular}
\end{scriptsize}
\end{table*}

\section{Data and Models}
\label{data-model-sec}

There is a long tradition in AI of using slightly synthetic tasks and
datasets in order to test the ability of particular models to achieve
these tasks \cite{weston2015towards,dodge2015evaluating}. The {\sc
  Personage} corpus \cite{oraby2018style} provides a controlled environment for testing
different models of neural generation and style generation. It
consists of 88,500 restaurant domain utterances whose style varies
according to models of personality, which were
generated by an existing statistical NLG engine that has the
capability of manipulating 67 different stylistic parameters
\cite{MairesseWalker10}.  Table \ref{table:multi-voice-examples} shows
sample utterances that are output for the singlevoice models and for
each of our multivoice models (described below) for the same MR.
Each output corresponding to each single voice personality is
controlled by a set of sentence planning parameters that vary for each
personality. These parameters are discussed in
Section~\ref{results-sec} when we evaluate stylistic fidelity.
What is important to note here is that each individual
voice represents a distinct stylistic distribution in the training
data.

The corpus uses the MRs and training/test splits of the E2E Generation
Challenge. There are 3,784 unique MRs in the training set, and the
corpus contains 17,771 MR/training utterance pairs for each of the
existing models for the personality traits of {\sc agreeable}, {\sc
  disagreeable}, {\sc conscientiousness}, {\sc unconscientiousness},
and {\sc extravert}, for a total training set of 88,855
utterances. This guarantees a wide range of variation in parameter
combinations.  The test set consists of 278 unique MRs. The
frequencies of longer utterances (more attribute MRs) vary across
train and test with test MRs not seen during training. The training
data has more smaller MRs, while the test set is more challenging, with more larger
MRs.


Previous work shows that a simple model trained on the whole corpus of
88,855 utterances produces semantically correct outputs, but
with reduced stylistic variation \cite{oraby2018style}, while a model
that allocates a variable corresponding to a
label for each style learns to reproduce the stylistic variation.
This is interesting because each style variable
(personality) actually encodes a set of 36 different stylistic
parameters and their values: the model learns for example how the {\sc
  disagreeable} personality tends to produce many shorter sentences in
the output, as well as learning that it tends to use expletives like
{\it damn}, e.g.  see the outputs based on {\sc disagreeable}
personality in Table~\ref{table:multi-voice-examples}.

\begin{figure}[t!h]
\centering
\includegraphics[width=2.3in]{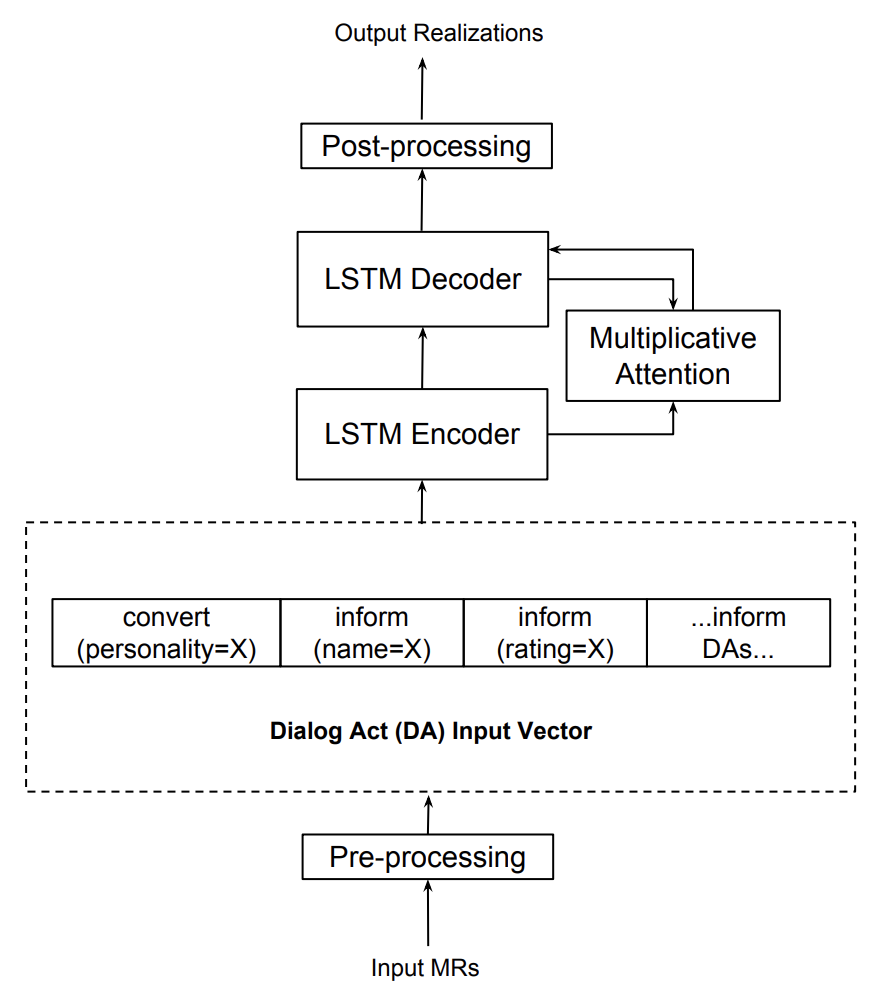}
\caption{\label{fig:model1}Neural network model architecture}
\end{figure}

\noindent{\bf Model Description.} Our {\sc nnlg} model uses a 
single token to represent personality encoding, following the use of single 
language labels used in machine translation and other
work on neural generation \cite{johnson2016google,oraby2018style}. Figure
\ref{fig:model1} summarizes the model architecture. This model 
builds on the open-source sequence-to-sequence (seq2seq) TGen system
\cite{DusekJ16a}, which is implemented in Tensorflow \cite{Abadi16}.\footnote{We refer 
the reader to TGen publications \cite{DusekJ16a, DusekJ16} for model details.}
The system is based on the seq2seq generation method with attention
\cite{DBLP:journals/corr/BahdanauCB14, sutskever2014sequence}, and
uses a sequence of LSTMs \cite{hochreiter1997long} for the encoder and
decoder, combined with beam-search and an n-best list reranker for
output tuning.

The inputs to the model are dialog acts for each system action (such
as {\it inform}) and a set of attribute slots (such as {\it rating})
and their values (such as {\it high} for attribute {\it rating}).  To
preprocess the corpus of MR/utterance pairs, attributes that take on
proper-noun values are delexicalized during training i.e. {\it name}
and {\it near}. We encode personality as an additional dialog act, of
type {\sc convert} with personality as the key and the target personality
as the value (see Figure~\ref{fig:model1}). For every input MR
and a personality, we train the model with the corresponding {\sc
  Personage} generated sentence.  Our model differs from the {\sc token} model used in our previous 
  work \cite{oraby2018style} because it is trained on unsorted inputs to
allow us to add multiple {\sc convert} tags to the MR at generation
time. Note that we do not train on multiple personalities, instead, we train one model that uses all the data, where each distinct
single personality has a corresponding {\sc convert(Personality = x)}
in the training instance.

At generation time, we generate singlevoice data for all the test
MRs (1,390 total realizations, 278 unique MRs, 
realized for each of 5 personalities).  For the multivoice
experiments, we generate 2 references per
combination of two personalities for each of the 278 test MRs, since
the order of the {\sc convert} tags matters.
For a given order, the model produces a single output. We
do not combine personalities that are exact opposites
such as {\sc agreeable} and  {\sc disagreeable}, yielding
8 combinations.  The multivoice test set consists of 4,448
total realizations (278 MRs and $8\times2$ outputs per MR).

\section{Results}
\label{results-sec}

Although it is well known that current automatic metrics do not perform
well for evaluating the quality of an NLG \cite{novikovanewnlg},
and that they penalize stylistic variation, we report automatic metrics
for completeness.  To address their limitations, we also report the
results of  our own metrics developed to measure semantic correctness and
stylistic fidelity. Examples of model outputs for single and multivoice are 
shown in Table \ref{table:multi-voice-examples}, demonstrating how our models  
interpolate the stylistic parameters described here.

\noindent {\bf Automatic Metrics.}  The automatic evaluation uses the
E2E generation challenge
script.\footnote{\url{https://github.com/tuetschek/e2e-metrics}}
Table~\ref{table:auto_metrics} summarizes the results for each
personality combination for the metrics: BLEU (n-gram precision), NIST
(weighted n-gram precision), METEOR (n-grams with synonym recall), and
ROUGE (n-gram recall). We note that multivoice automatically has a
better chance because the evaluation is over 4,448 examples as opposed
to 1,390 for singlevoice, and each multivoice output is compared to 2
possible references (one for each single voice), and then averaged.

\begin{table}[!htb]
\centering 
\caption{\label{table:auto_metrics} Automatic metric evaluation}
\begin{footnotesize}
\begin{tabular}
{@{} p{0.75in}|p{0.38in}|p{0.38in}|p{0.38in}|p{0.38in} @{}}
\hline
{\bf \scriptsize Personality} & {\bf \scriptsize BLEU} & {\bf \scriptsize  NIST} & {\bf \scriptsize METEOR} & {\bf \scriptsize ROUGE\_L} \\ \hline     
{\sc \scriptsize SingleVoice } & 0.35  & 4.93 & 0.36  & 0.50\\                            
{\sc \scriptsize MultiVoice} & 0.42  & 5.64 & 0.36  & 0.52 \\\hline
\end{tabular}
\end{footnotesize}
\end{table}

\noindent{\bf Semantic Errors.} Table \ref{table:deletes-repeats} shows ratios for the number of
deletions, repeats, and hallucinations for each single and multivoice
model for their respective test sets (1,390 total realizations and
4,448 realizations).  The error counts are split by personality, and
normalized by the number of unique MRs (278). Note that smaller ratios
are preferable, indicating fewer errors.  As we predicted, it is more
challenging to preserve semantic fidelity when attempting to hit
multiple stylistic targets. We see that in most cases the frequency of
errors increase for multivoice compared to singlevoice, with
particular combinations such as {\sc disagreeable} plus {\sc
  extraversion} making more than one attribute deletion for each
output on average.  In the singlevoice results {\sc disagreeable} and
{\sc extravert} make the most errors with the smallest total ratio
found for {\sc conscientious}, but when {\sc conscientious} combines
with {\sc disagreeble} it performs worse than either model alone.

\begin{table}[!htb]
\centering 
\caption{\label{table:deletes-repeats} Ratio of errors by multivoice personality pairs as compared to singlevoice models}
\vspace{-.1in}
\begin{footnotesize}
\begin{tabular}
{@{} p{0.85in}|p{0.45in}|p{0.45in}|p{0.6in} @{}}
\hline
{\bf \scriptsize Personality} & {\bf \scriptsize Deletions} & {\bf \scriptsize  Repetitions} & {\bf \scriptsize Hallucinations}  \\ \hline\hline  
{\sc \scriptsize Agree } & 0.27  & 0.29 & 0.34  \\
{\sc \scriptsize Consc } & 0.22  & 0.12 & 0.41  \\
{\sc \scriptsize Extra } & 0.74 & 0.46 & 0.35  \\
{\sc \scriptsize UnConsc } & 0.31  & 0.28 & 0.29  \\
{\sc \scriptsize Disagree } & 0.87 & 0.81 & 0.22  \\ \hline 
{\bf \scriptsize Personality Pairs} & & &   \\ \hline\hline  
{\sc \scriptsize Agree+Consc } & 0.44  & 0.08 & 0.26  \\                            
{\sc \scriptsize Agree+Extra } & 0.28  & 0.17 & 0.19  \\
{\sc \scriptsize Agree+Unconsc } & 0.33 & 0.24 & 0.24  \\  
{\sc \scriptsize Consc+Disagr } & 1.01  & 0.18 & 0.28  \\                            
{\sc \scriptsize Consc+Extra } & 0.67  & 0.28 & 0.23  \\
{\sc \scriptsize Disagr+Extra } & 1.20 & 0.75 & 0.09  \\
{\sc \scriptsize Disagr+Unconsc } & 1.10  & 0.39 & 0.14  \\                            
{\sc \scriptsize Extra+Unconsc } &  1.05 & 0.55 & 0.17 \\\hline
\end{tabular}
\end{footnotesize}
\end{table}

\noindent{\bf Stylistic Characterization.} To
characterize the differences in style between the multivoice and
singlevoice outputs, we develop scripts that count the aggregation
operations and pragmatic markers in Figure~\ref{table:agg-prag} in
both the singlevoice and multivoice test data.  We then compare
the singlevoice data directly with multivoice results.

\begin{table}[htb!]
\centering \caption{\label{table:agg-prag} Aggregation and Pragmatic Operations}
\vspace{-.1in}
\begin{footnotesize}
\begin{tabular}
{@{} p{1.2in}|p{1.5in} @{}}
\hline
{\bf Attribute} & {\bf Example} \\ \hline\hline
\multicolumn{2}{l}{ \cellcolor[gray]{0.9} {\sc Aggregation Operations}}    \\               
{\sc Period} &  {\it X serves Y. It is in  Z.}  \\
{\sc "With" cue} &  {\it X is in Y, with Z.}  \\
{\sc Conjunction} & {\it X is Y and it is Z. \& X is Y, it is Z.} \\
{\sc All Merge} & {\it X is Y, W and Z \& X is Y in Z}  \\
{\sc "Also" cue} &  {\it X has Y, also it has Z.} \\
\multicolumn{2}{l}{ \cellcolor[gray]{0.9} {\sc Pragmatic Markers}}     \\    
{\sc ack\_definitive} & \it right, ok \\ 
{\sc ack\_justification} & \it I see, well \\ 
{\sc ack\_yeah} & \it yeah\\ 
{\sc confirmation} & \it let's see what we can find on X, let's see ....., did you say X?  \\ 
{\sc initial rejection} & \it mmm, I'm not sure, I don't know.  \\ 
{\sc competence mit.} & \it come on, obviously, everybody knows that \\ 
{\sc filled pause stative} & \it err, I mean, mmhm \\ 
{\sc down\_kind\_of} & \it kind of \\ 
{\sc down\_like} & \it like \\ 
{\sc down\_around} & \it around \\  
{\sc exclaim} & \it ! \\
{\sc indicate surprise} & \it oh \\
{\sc general softener} & \it sort of, somewhat, quite, rather \\ 
{\sc down\_subord} & \it I think that, I guess \\ 
{\sc emphasizer} & \it really, basically, actually, just \\ 
{\sc emph\_you\_know} & \it you know \\ 
{\sc expletives} \& & \it oh god, damn, oh gosh, darn  \\ 
{\sc in group marker} & \it pal, mate, buddy, friend \\ 
{\sc tag question} & \it alright?, you see? ok? \\ 
\hline
\end{tabular}
\end{footnotesize}
\end{table}

The aggregation parameters in Table~\ref{table:agg-prag} control how
the NLG combines attributes into sentences, e.g. whether it tries to
create complex sentences and what
types of combination operations it uses.  The pragmatic operators in
the bottom part of Table~\ref{table:agg-prag} are intended to achieve
particular pragmatic effects in the generated outputs: for example the
use of a hedge such as {\it sort of} softens a claim and affects
perceptions of friendliness and politeness \cite{BrownLevinson87},
while the exaggeration associated with emphasizers like {\it actually,
  basically, really} influences perceptions of extraversion and
enthusiasm \cite{OberlanderGill04b,DewaeleFurnham99}. Each
  parameter value can be set to {\tt high}, {\tt low}, or {\tt don't
    care}.

\noindent {\bf Aggregation.}\label{eval:aggreg} To measure the
similarity of each multivoice model to its parent single voices for
aggregation operations, we first count the average number of times
each aggregation operation occurs for each model and personality or
personality combination. We then compute Pearson correlation across
different model outputs to quantify the similarity of these model
outputs with respect to the aggregation operations. Table
\ref{table:aggreg-results} provides a summary of these results (higher means 
more correlated).

The final column of Table \ref{table:aggreg-results} provides the
correlations between the original two single voices that were put
together to create the multivoice model. This shows for example (Row
1) that {\sc agreeable} and {\sc conscientious} are similar in their
use of aggregation but that {\sc disagreeable} and {\sc extraversion}
are very dissimilar (Row 6).  We would expect that models that are
similar to start with would be less novel when they are combined, and
indeed Row 1 shows that when the multivoice model is compared with
both the original {\sc agreeable} voice (Column 3) and the {\sc
  conscientious} voice (Column 4) the use of aggregation operations
changes little. However other combinations seem to produce completely
novel models that use aggregation very differently than either of
their singlevoice source models. For example in Row 7 the combination
of {\sc disagreeable} and {\sc unconscientious} produces a model whose
use of aggregation is distinct from either of its source models.  All
of the correlations in Table \ref{table:aggreg-results} are
significant ($p<0.05$) except for the 0.01 correlation when comparing
the single voices of {\sc conscientious} vs. {\sc disagree} where the
p-value is 0.6.

\begin{table}[!htb]
\centering \caption{\label{table:aggreg-results} Correlations between {\sc personage}  data and multivoice models for the aggregation operations in Table \ref{table:agg-prag}}
\begin{footnotesize}
\begin{tabular}
{@{} p{0.1in}|p{0.3in}|p{0.36in}|r|r|r @{}}
\hline
{\bf \scriptsize \#} & {\bf \scriptsize P1} &{\bf \scriptsize P2} & {\bf \scriptsize P1+P2 vs. P1} & {\bf \scriptsize  P1+P2 vs. P2} & {\bf \scriptsize P1 vs. P2}  \\ \hline\hline  
1 & {\sc \scriptsize Agree} &{\sc \scriptsize Consc } & 0.74  & 0.76 & 0.74  \\                            
2 & {\sc \scriptsize Agree} &{\sc \scriptsize Extra } & 0.70  & 0.31 & 0.44  \\
3 & {\sc \scriptsize Agree} &{\sc \scriptsize Unconsc }& 0.75 & 0.31 & 0.65  \\  
4 & {\sc \scriptsize Consc} &{\sc \scriptsize Disagr } & 0.36  & 0.65 & 0.01  \\                            
5 & {\sc \scriptsize Consc} &{\sc \scriptsize Extra } & 0.51  & 0.31 & 0.44  \\
6 & {\sc \scriptsize Disagr} &{\sc \scriptsize Extra } & 0.53 & -0.36 & -0.04  \\
7 & {\sc \scriptsize Disagr} &{\sc \scriptsize Unconsc } & 0.23  & 0.33 & 0.05  \\                            
8 & {\sc \scriptsize Extra} &{\sc \scriptsize Unconsc } &  0.20 & 0.43 & 0.47 \\\hline
\end{tabular}
\end{footnotesize}
\vspace{-.1in}

\end{table}

\begin{figure}[ht!b]
    \centering
\includegraphics[width=2.0in]{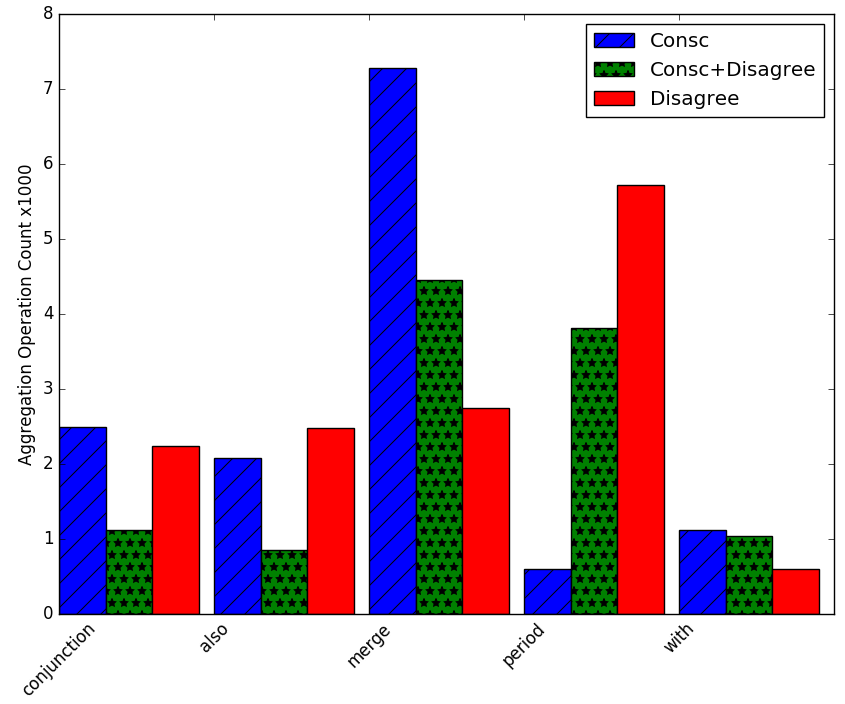}
\vspace{-.1in}        
    \caption{Frequency of the most frequent aggregation operations for Conscientiousness and Disagreeable 
compared to combined Conscientiousness and Disagreeable multivoice   \label{fig:agg_plot}}
\end{figure}

Figure~\ref{fig:agg_plot}  provides a closer look at particular
aggregation operations associated with {\sc conscientiousness} and {\sc
  disagreeable} and plots the differences between the singlevoice
models and the use of these operations in the multivoice
models. Interestingly, these plots also clearly show that the multivoice
model is a novel personality, yielding a different
distribution for aggregation operations than either of its source voice
styles.

\noindent {\bf Pragmatic Marker Usage.}
To measure the models' use of pragmatic markers, we count the number
of times each  marker in Table
\ref{table:agg-prag} occurred in the model outputs, compared
to the singlevoice references. We again  compute the
Pearson correlation between the original voices and the multivoice  model outputs for
personality combination. The results are shown in 
Table \ref{table:prag-results} (all correlations significant with $p \le 0.05$). 

\begin{table}[!htb]
\centering \caption{\label{table:prag-results} Correlations between {\sc personage} data and multivoice models for the pragmatic markers in Table \ref{table:agg-prag}}
\begin{footnotesize}
\begin{tabular}
{@{} p{0.1in}|p{0.3in}|p{0.36in}|r|r|r @{}}
\hline
{\bf \scriptsize \#} & {\bf \scriptsize P1} &{\bf \scriptsize P2} & {\bf \scriptsize P1+P2 vs. P1} & {\bf \scriptsize  P1+P2 vs. P2} & {\bf \scriptsize P1 vs. P2}  \\ \hline\hline  
1 & {\sc \scriptsize Agree} &{\sc \scriptsize Consc } & 0.11  & 0.74 & 0.30  \\                            
2 & {\sc \scriptsize Agree} &{\sc \scriptsize Extra } & 0.19  & -0.02 & -0.07  \\
3 & {\sc \scriptsize Agree} &{\sc \scriptsize Unconsc }& 0.03 & 0.18 & -0.16  \\  
4 & {\sc \scriptsize Consc} &{\sc \scriptsize Disagr } & 0.44  & 0.05 & -0.10  \\                            
5 & {\sc \scriptsize Consc} &{\sc \scriptsize Extra } & 0.41  & -0.09 & -0.11  \\
6 & {\sc \scriptsize Disagr} &{\sc \scriptsize Extra } & 0.12 & -0.03 & -0.07  \\
7 & {\sc \scriptsize Disagr} &{\sc \scriptsize Unconsc } & 0.09  & 0.34 & -0.05  \\                            
8 & {\sc \scriptsize Extra} &{\sc \scriptsize Unconsc } &  -0.11 & 0.37 & -0.08 \\\hline
\end{tabular}
\end{footnotesize}
\vspace{-.1in}
\end{table}

The final column of Table \ref{table:prag-results} provides the
correlations between the original two single voices that were put
together to create the multivoice model. As we can see in
Row 1, the only two voices that are similar to start
are {\sc agreeable} and {\sc conscientious}. All of the other voices have
negative correlations with one another in their use of pragmatic markers.
Interestingly, the multivoice combination of {\sc agreeable} and
{\sc conscientious} resembles {\sc conscientious} much more (see column 4).
All the other multivoice models also appear to resemble one of the parent models
more than the other, but none  are {\it very} similar to their
parents: they each appear to demonstrate characteristics of a novel voice. 
For example, in Row 6, the combination
of {\sc disagreeable} and {\sc extraversion} produces a model that
bears very little similarity to either {\sc disagreeable} (0.12 correlation)
or {\sc extraversion} (-0.03 correlation).

\begin{figure}[ht!b]
    \centering
\vspace{-.1in}        
     \includegraphics[width=2.0in]{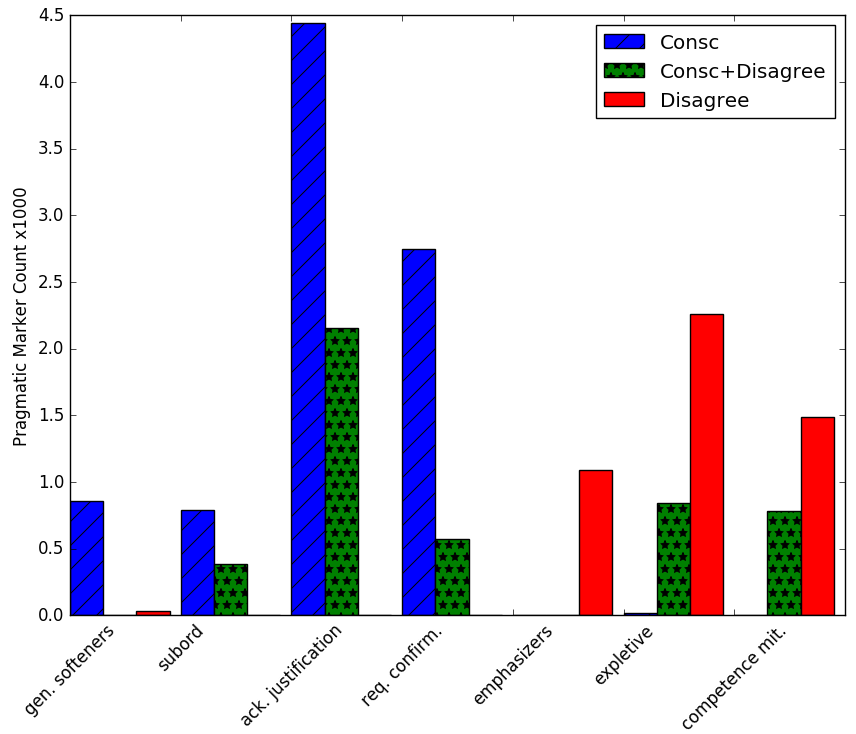}
    \caption{Frequency of the most frequent pragmatic markers for Conscientiousness and Disagreeable 
compared to combined Conscientiousness and Disagreeable multivoice \label{fig:prag_plot}}

\end{figure}

Figure~\ref{fig:prag_plot} provides a closer look at particular
pragmatic markers associated with {\sc conscientiousness} and {\sc
  disagreeable} and plots the differences between the singlevoice
models and the  multivoice
models. Again, interestingly, these plots show that the multivoice
model is a novel personality that yields a different
distribution for pragmatic markers than either of its source voice
styles.

\section{Related Work and Conclusion}
\label{rel-conc-sec}

The restaurant domain has been a testbed for conversational agents for
over 25 years
\cite{PHSZ92,StentPrasadWalker04,devillers2004french,gavsic2008training,MairesseBagel},
but there is little previous work examining stylistic variation in
this domain \cite{MairesseWalker10,dethlefs2014cluster}.  Most of the
recent research using neural NLG has focused on semantic fidelity
\cite{Wenetal15,Mei2015,DusekJ16,Lampouras2016}, however there is work
on methods for controlling when long utterances should be split into
shorter ones, and for attempting to enforce pronominalization
\cite{Nayaketal17}.  Other work has pointed out how poor evaluation
metrics such as BLEU are for evaluating natural language generation
quality \cite{novikovanewnlg}.

Recent work on neural methods for controlling linguistic style has mainly 
been carried out in the context of machine translation
\cite{sennrich2016controlling} or focused on tasks where semantic
fidelity was not required \cite{Ficler17}.  Previous work
in the statistical NLG tradition presents methods for controlling stylistic
variation \cite{PaivaEvans2004,Isardetal06,InkpenHirst04}.  Work on the
{\it persona} of a conversational agent did not actually focus on
stylistic variation, or personality, but instead tried to ensure that
an open domain conversational agent would answer questions about
itself in a semantically consistent way \cite{li2016persona}.

Here we present the first experiment, to our knowledge, examining
stylistic generalization in a domain that requires semantic fidelity.
We show that our neural models produce novel styles that they have
not seen in training, and examine how and to what extent stylistic control
interacts with semantic fidelity.

\bibliography{phd,nl,nlg,naaclhlt2018,../../nl}
\bibliographystyle{IEEEtran}

\end{document}